\documentclass{article}
\usepackage{spconf,amsmath,graphicx}
\usepackage{framed,multirow}
\usepackage{amssymb}
\usepackage{latexsym}
\usepackage{amsmath}
\usepackage{url}
\usepackage{xcolor}
\definecolor{newcolor}{rgb}{.8,.349,.1}
\usepackage[super]{nth}
\usepackage{tabularx}
\usepackage{subcaption}
\newcommand{\etal}{\textit{et al.}}
\newcommand{\Skip}[1]{}

\title{Enhancing Visual Re-ranking through \\ Denoising Nearest Neighbor Graph via Continuous CRF}

\name{Jaeyoon Kim, Yoonki Cho, Taeyoung Kim\textsuperscript{*}\thanks{\textsuperscript{*}Currently at Samsung Research, Samsung Electronics Co., Ltd.}, and Sung-Eui Yoon}
\address{KAIST}

\begin{document}
\maketitle
\begin{abstract}
Nearest neighbor~(NN) graph based visual re-ranking has emerged as a powerful approach for improving retrieval accuracy, offering the advantages of effectively exploring high-dimensional manifolds without requiring additional fine-tuning. 
However, the effectiveness of NN graph-based re-ranking is fundamentally constrained by the quality of its edge connectivity, as incorrect connections between dissimilar (negative) images frequently occur. 
This is known as a \textit{noisy edge} problem, which hinders the re-ranking performance of existing techniques and limits their potential.
To remedy this issue, we propose a \textit{complementary} denoising method based on Continuous Conditional Random Fields (C-CRF) that leverages statistical distances derived from similarity-based distributions. 
As a pre-processing step for enhancing NN graph-based retrieval, our approach constructs fully connected cliques around each anchor image and employs a novel statistical distance metric to robustly alleviate noisy edges before re-ranking while achieving efficient processing through offline computation.
Extensive experimental results demonstrate that our method consistently improves three different NN graph-based re-ranking approaches, yielding significant gains in retrieval accuracy.
\end{abstract}


\vspace{-2mm}
\section{Introduction}
\vspace{-2mm}
\label{sec1}
\begin{figure}[t]
	\centering
	\begin{subfigure}[b]{0.21\textwidth}
		\captionsetup{justification=centering}
		\includegraphics[width = 0.85\textwidth]{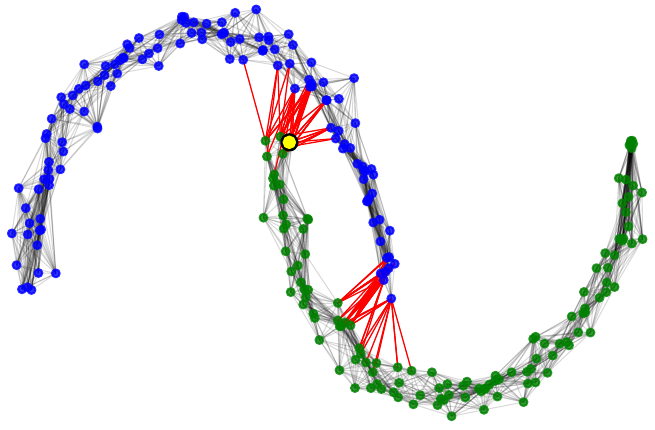}
		\caption{Initial NN graph}
	\end{subfigure}%

        \hfill
	\begin{subfigure}[b]{0.21\textwidth}
		\captionsetup{justification=centering}
		\includegraphics[width = 0.85\textwidth]{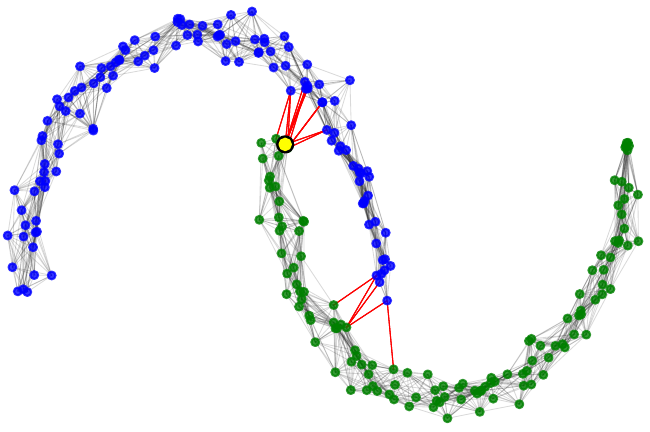}
		\caption{NN graph after \\ reciprocity check}
	\end{subfigure}%
	\hspace{2mm}
	\begin{subfigure}[b]{0.21\textwidth}
		\captionsetup{justification=centering}
		\includegraphics[width = 0.85\textwidth]{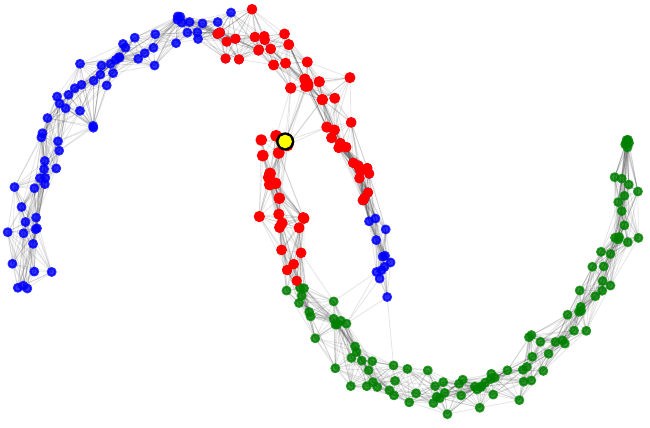}
		\caption{Diffusion-based \\ retrieval of (b)}
	\end{subfigure}
        \hfill
        
        \hfill
	\begin{subfigure}[b]{0.21\textwidth}
		\captionsetup{justification=centering}
		\includegraphics[width = 0.85\textwidth]{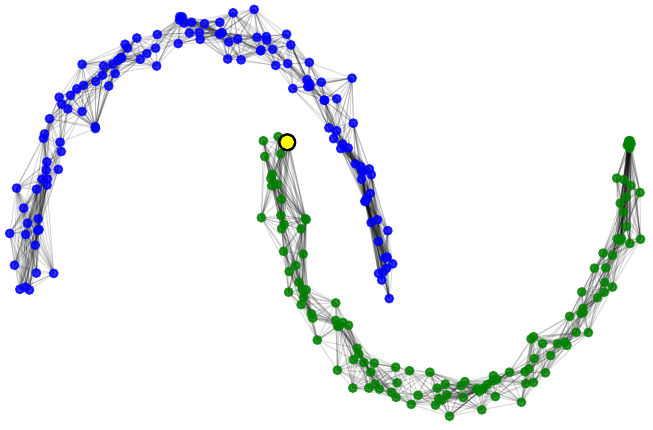}
		\caption{NN graph denoised by \\ our method}
	\end{subfigure}%
	\hspace{2mm}
	\begin{subfigure}[b]{0.21\textwidth}
		\captionsetup{justification=centering}
		\includegraphics[width = 0.85\textwidth]{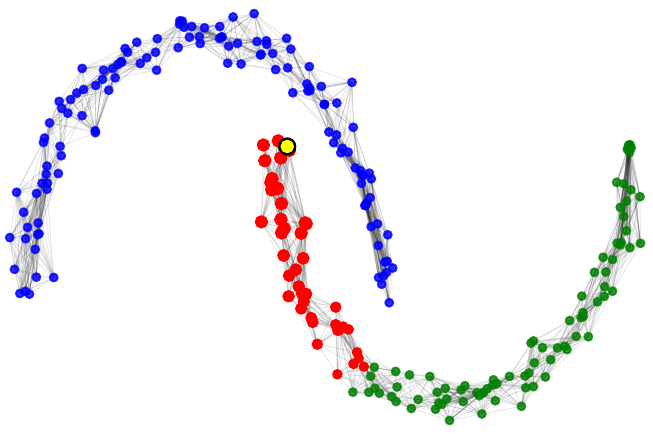}
		\caption{Diffusion-based \\ retrieval of (d)}
	\end{subfigure}
        \hfill
        
         \vspace{-3mm}	
	\caption{
            This figure shows NN graph (affinity) of toy data on manifold.
            The blue and green colors of nodes indicate different manifolds, while a yellow node and red nodes represent a query point and the retrieval results from a visual re-    ranking method of diffusion.
            The red edges are noisy edges that are connected between different manifolds. Results (e) of our approach (d) are on the correct manifold.
	}
	\label{fig:toy_result}
    \vspace{-7mm}
\end{figure}

Visual retrieval systems typically perform nearest neighbor search with visual descriptors; however, they face fundamental challenges when handling non-linear manifold structures in high-dimensional feature spaces~\cite{gordo2016deep, zhou2017recent}.
To address this issue, re-ranking process has emerged as a crucial component in visual retrieval systems, refining initial search results to improve retrieval accuracy.
By leveraging various contextual information and relationships between images, re-ranking methods have consistently demonstrated their ability to enhance the quality of retrieval results beyond the initial ranking~\cite{DBLP:journals/pami/BaiBTL19, philbin2007object, arandjelovic2012three}.

Visual re-ranking techniques have evolved along several distinct research directions.
Geometric verification approaches~\cite{philbin2007object, simeoni2019local, an2023topological} leverage local descriptors to verify spatial consistency between image pairs, effectively reducing false matches but incurring significant computational overhead.
In contrast, query expansion methods~\cite{DBLP:conf/iccv/ChumPSIZ07, arandjelovic2012three, DBLP:journals/pami/RadenovicTC19} utilize global descriptors to explore image manifolds, aggregating nearest neighbor features to refine the original query descriptor for subsequent retrieval processes.
Learning-based approaches~\cite{gordo2020attention, ouyang2021contextual} employ dedicated re-ranking models with neural networks, however, these methods typically require additional fine-tuning or training data.
Meanwhile, nearest neighbor graph-based methods~\cite{DBLP:conf/nips/ZhouWGBS03, DBLP:conf/cvpr/DonoserB13} have gained prominence for their ability to explore high-dimensional manifolds without requiring additional training.
These approaches, particularly diffusion processes~\cite{DBLP:conf/cvpr/IscenTAFC17, DBLP:journals/pami/BaiBTL19} and graph traversal methods~\cite{chang2019explore, DBLP:journals/spic/QiL16}, achieve superior performance by effectively utilizing comprehensive affinity relationships within the image database.

However, the effectiveness of nearest neighbor graph-based re-ranking is fundamentally constrained by the quality of its edge connectivity in the graph structure.
Specifically, incorrect connections between dissimilar (negative) images frequently occur in the initial nearest neighbor graph, known as the \textit{noisy edge} problem.
Such noisy edges significantly hinder re-ranking performance by introducing false connectivity paths that propagate incorrect similarity information throughout the manifold structure.
While previous works have attempted to address this issue through locally constrained affinity construction~\cite{DBLP:conf/cvpr/YangKL09} and reciprocity checks~\cite{DBLP:conf/cvpr/IscenTAFC17}, these solutions are limited as they only consider pairwise relationships between edges.
The reciprocity check, in particular, fails to fully exploit collective information across the graph, as it only considers mutual neighbor relationships while ignoring broader neighborhood structures, leading to suboptimal performance in complex manifold scenarios.

To overcome these limitations, we propose a \textit{complementary} denoising approach based on Continuous Conditional Random Fields (C-CRF)~\cite{sutton2007introduction} that leverages statistical distances derived from similarity-based distributions.
While CRF has been extensively studied for error reduction in various computer vision tasks such as segmentation and depth estimation~\cite{chen2017deeplab, DBLP:conf/nips/QinLZWL08, DBLP:conf/aaai/RistovskiRVO13}, its potential for addressing the noisy edge problem in visual retrieval remains unexplored.
As a pre-processing step for enhancing NN graph-based retrieval, our method constructs fully connected cliques around each anchor image and considers relationships among all edges within these cliques to reach a more robust consensus, effectively addressing the limitations of pairwise verification approaches.
Unlike previous applications of CRF, our approach uniquely focuses on refining pairwise similarities between database images through efficient offline processing, making them robust against noisy edges and thereby improving re-ranking performance.
Extensive experimental results demonstrate that our method consistently enhances three different nearest neighbor graph-based re-ranking approaches, yielding significant improvements in retrieval accuracy.

In summary, our contributions are as follows:
\begin{itemize}
    \vspace{-1mm}
    \item We propose a complementary denoising method based on C-CRF that enhances NN graph-based re-ranking without requiring additional networks or fine-tuning.
    \item We design a statistical distance metric derived from similarity-based distributions and employ fully connected cliques for efficient offline graph denoising.
    \item Experimental results demonstrate that our method consistently improves three different NN graph-based re-ranking approaches, confirming its effectiveness as an easily integrable complementary tool.
\end{itemize}

\vspace{-2mm}
\section{Preliminaries}
\vspace{-2mm}
In this section, we summarize the diffusion process to show how the NN-graph can be applied for visual re-ranking.  
The diffusion process in visual re-ranking constructs revised similarity considering manifold from initial similarity computed from all images in the
database.
Generally, this process constructs an undirected graph structure from the affinity matrix, which is defined using the pairwise similarity between images, and performs diffusion on this graph using a random walk formulation~\cite{DBLP:conf/cvpr/IscenTAFC17,DBLP:journals/corr/abs-1811-10907}.

We define a database $\mathcal{X}$ as a set of $N$ image descriptors, where each $\mathbf{x}_i$ denotes a descriptor of image $i$ with dimensionality $d$; $\mathbf{x}_i \in \mathbb{R}^d$.  We define the affinity matrix $\mathbf{A} = (a_{ij}) \in \mathbb{R}^{N \times N}$, $\forall i,j ~\in~ \{1,\dotsc,N\}$, following the reciprocity check with a local constraint by considering $k$ nearest neighbors~\cite{DBLP:conf/cvpr/IscenTAFC17}, where each element is obtained by:

\begin{equation}
	s_{k} (\mathbf{x}_i|\mathbf{x}_j) =
	\begin{cases}
		s(\mathbf{x}_i, \mathbf{x}_j)      & \mathbf{x}_i \in \mathrm{NN}_k (\mathbf{x}_j)\\
		0  & \mathrm{otherwise}
	\end{cases},
\end{equation}

\begin{equation}
	a_{ij} = \min\{s_{k}(\mathbf{x}_i|\mathbf{x}_j), s_{k}(\mathbf{x}_j|\mathbf{x}_i)\},
	\label{eq:reciprocity}
\end{equation}
where the similarity function $s(\cdot,\cdot)$ is positive and has zero self-similarity, and $\mathrm{NN}_k(\mathbf{x})$ denotes $k$-NN of $\mathbf{x}$.
Intuitively, $a_{ij}$ equals to $s(\mathbf{x}_i, \mathbf{x}_j) $ if $\mathbf{x_i}$, $\mathbf{x_j}$ are the $k$ nearest neighbors of each other, and zero otherwise.

The degree matrix $\mathbf{D}$ is a diagonal matrix with the row-wise sum of $\mathbf{A}$.
It is used to symmetrically normalize $\mathbf{A}$ as: $\mathbf{S}= \mathbf{D}^{-1/2}\mathbf{A}\mathbf{D}^{-1/2}$.
After the matrix computation, starting from an arbitrary vector $\mathbf{v}^0$ that represents the initial similarity of a given query $\mathbf{y}$, the diffusion is performed until its state converges with a random walk iteration:
\begin{equation}
	\mathbf{v}^{t+1} = \rho \mathbf{Sv}^{t} + (1-\rho)\mathbf{v}^0, ~ \rho \in (0, 1).
\end{equation}
Zhou~\etal~\cite{DBLP:conf/nips/ZhouWGBS03} shows this iteration converges to a closed-form solution when assuming $0<\rho<1$:
\begin{equation}
	\mathbf{v}^* = (1-\rho)(\mathbf{I} - \rho \mathbf{S})^{-1} \mathbf{v}^0.
\end{equation}
In the conventional diffusion process, the values in $\mathbf{f}^*$ contain the refined similarity of the given query $\mathbf{y}$. Generally, the initial state $\mathbf{f}^0$ represents the similarities between the query $\mathbf{y}$ and the corresponding $k$ nearest neighbors as:
\begin{equation}
	\mathbf{v}^0 = s_k (\mathbf{x}_i | \mathbf{y}),~~ \forall i \in \{1, \dots,N\}.
	\label{Eq:new_query}
\end{equation}

In this paper, we propose a novel, denoising approach for the initial affinity matrix to reduce the noisy edges. 
\vspace{-2mm}
\section{Approach}
\vspace{-2mm}
Diffusion methods~\cite{DBLP:conf/cvpr/IscenTAFC17,DBLP:journals/corr/abs-1811-10907} for image retrieval adopted a simple approach of using the reciprocity check (Eq.~\ref{eq:reciprocity}) with $k$-NN of each database image to ameliorate noisy edges whose vertices are not positive, i.e., similar images.
We, however, found that the reciprocity check is limited to handling noise, since identifying the noisy edge is based on the consensus of only two nodes of an edge, resulting incorrect identification of noise edges.

\begin{figure}[t]
	\centering
	\begin{subfigure}[b]{0.13\textwidth}
		\captionsetup{justification=centering}
		\includegraphics[height = 0.13\textheight]{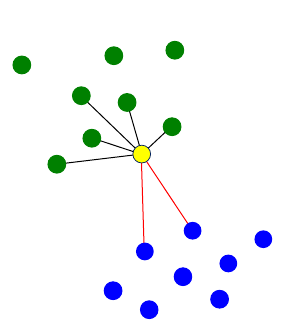}
		\caption{}
	\end{subfigure}%
	\hfill
	\begin{subfigure}[b]{0.13\textwidth}
		\captionsetup{justification=centering}
		\includegraphics[height = 0.13\textheight]{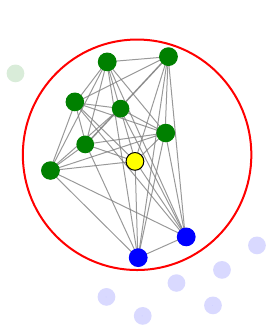}
		\caption{}
	\end{subfigure}%
	\hfill
	\begin{subfigure}[b]{0.13\textwidth}
		\captionsetup{justification=centering}
		\includegraphics[height = 0.13\textheight]{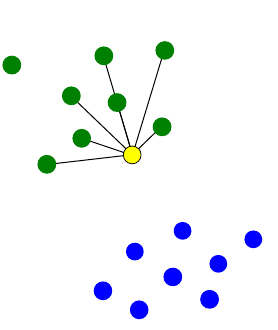}
		\caption{}
	\end{subfigure}%
	\vspace{-3mm}
	\caption{An example of clique-based denoising; $k=7$ and $L=10$. (a) shows initial edges of a yellow pivot node, and red edges are noisy edges. 
		(b) shows a clique of the pivot, and all nodes in the clique are fully connected. 
		(c) represents edges denoised by C-CRF, which is constructed from the clique. }
	\label{fig:overview_method}\vspace{-5mm}
\end{figure}

While this noisy edge problem was not seriously treated in diffusion approaches for image retrieval, we tackle the noisy edge problem  by considering all neighbors from each image in a well-connected subgraph where we can robustly detect noisy edges. Specifically, we realize our goal by using C-CRF on subgraphs, i.e., cliques, with our proposed weight function~(illustrated in Fig.~\ref{fig:overview_method}).

\vspace{-2mm}
\subsection{C-CRF on clique}
\vspace{-2mm}

Our goal is to refine initial similarities of NN graph that are commonly set by top-$k$ similarities of each database image.  
For computational efficiency and effective noise handling, we utilize the concept of cliques rather than processing the entire graph. Each clique is then treated with C-CRF to optimize performance.
Specifically, we choose a clique, a complete subgraph, where we can have ample information such that we can identify noisy edges.

We suppose that the whole database size is $N$ and a clique of a pivot image $\left\{ {{I_p}} \right\}_{p = 1}^N$ is ${{\cal C}_p} = \left(
{{{\cal N}_p},{{\cal E}_p}} \right)$, where a set of nodes (images) ${\cal
	N}_p$ is composed of $L$ nearest neighbors from the pivot image $I_p$ 
by a specified clique size $L$, and ${\cal E}_p$ is a set of edges that makes the clique ${\cal C}_p$, i.e., a complete subgraph.
Given a clique ${\cal C}_p$, its set of initial similarities can be computed by the cosine similarity or Euclidean distance. 
For simplicity, we express $s(\mathbf{x}_i, \mathbf{x}_j)$ as ${s_{i,j}}$.  
The set is fed as input to C-CRF and is defined:
${{\bf{S}}_{{{\cal C}_p}}} = \left\{ {{s_{i,j}}| i \in {{\cal N}_p},j \in {{\cal N}_p}} \right\}$
with the given initial similarity $s_{i,j}$ on clique ${\cal C}_p$.
C-CRF computes a set of refined similarities defined as ${{\bf{y}}_p} = \left\{ {{y_{p,i}}| i \in {{\cal N}_p}} \right\}$, which has $L$ elements with refined similarity $y_{p,i}$.

C-CRF on each clique ${\cal C}_p$ is represented by a conditional probability as follows:
\begin{equation}
	\begin{array}{l}
		{{\rm{P}}_p}\left( {{{\bf{y}}_p}|{{\bf{S}}_{{{\cal C}_p}}}} \right) = \frac{1}{{Z\left( {{{\bf{S}}_{{{\cal C}_p}}}} \right)}}\exp \left\{ { - E\left( {{{\bf{y}}_p},{{\bf{S}}_{{{\cal C}_p}}}} \right)} \right\},
	\end{array}
	\label{Eq:7}
\end{equation} 
where $Z\left( {{{\bf{S}}_{{{\cal C}_p}}}} \right) = \int_{{\bf{y}}_p} {\exp \left\{ { - E\left( {{{\bf{y}}_p},{{\bf{S}}_{{{\cal C}_p}}}} \right)} \right\}d{\bf{y}}_p}$ is a normalization constant.
The corresponding energy function, $E\left( {{{\bf{y}}_p},{{\bf{S}}_{{{\cal C}_p}}}} \right)$, can be defined by linearly combining a unary potential $\phi \left( \cdot \right)$ and a pairwise potential $\psi \left( \cdot \right) $ with positive scalars $\alpha$ and $\beta$, as follows:
\begin{equation}
	\begin{array}{l}
		\alpha\sum\nolimits_{i \in {{\cal N}_p}} {\phi \left( {{y_{p,i}},{{\bf{S}}_{{{\cal C}_p}}}} \right) + \frac{\beta}{2}\sum\nolimits_{\left( {i,j} \right) \in {{\cal E}_p}} {\psi \left( {{y_{p,i}},{y_{p,j}},{{\bf{S}}_{{{\cal C}_p}}}} \right)} }. 
	\end{array}
\end{equation}
The unary potential, $\phi \left( {{y_{p,i}},{{\bf{S}}_{{{\cal C}_p}}}} \right) = {\left( {{y_{p,i}} - {s_{p,i}}} \right)^2}$, intends $y_{p,i}$ to follow the initial similarity $s_{p,i}$.
To consider inter-image relationships, the pairwise potential is defined as:
\begin{equation}
	\psi \left( {{y_{p,i}},{y_{p,j}},{{\bf{S}}_{{{\cal C}_p}}}} \right) =  \textrm{w}({{\bf{f}}_i},{{\bf{f}}_j},{{\bf{S}}_{{{\cal C}_p}}}){\left( {{y_{p,i}} - {y_{p,j}}} \right)^2},
\end{equation}
where ${\bf{f}}_i$ and ${\bf{f}}_j$ are feature vectors of node $i$ and node $j$, respectively, and weight $\textrm{w}({{\bf{f}}_i},{{\bf{f}}_j},{{\bf{S}}_{{{\cal C}_p}}})$ is introduced to measure a conformity between node $i$ and node $j$ in the feature space and similarity-based distribution.
Intuitively, the pairwise potential encourages that  as a higher weight $\textrm{w}({{\bf{f}}_i},{{\bf{f}}_j},{{\bf{S}}_{{{\cal C}_p}}})$ of node $i$ and node $j$, the closer refined similarity values between $y_{p,i}$ and $y_{p,j}$.
These potentials are designed to reach a consensus from  all the nodes on a clique for refining the initial similarities.

While we perform C-CRF on cliques to be robust against noisy edges, we further found that, in challenging datasets, a high number of noisy edges can exist even
in the clique, deteriorating the similarity refinement process through C-CRFs.
As a result, we design the weight function to be robust even in this extreme case by utilizing a statistical distance considering a similarity-based  distribution
in the clique.

More specifically, our probability mass function (PMF) in terms of a node $i$ is defined by $l_2$-normalization within a clique of a pivot node $p$, followed 
by a softmax as:
\begin{equation}
	{Q_{i}}\left( k \right) = \frac{{\exp \left( {{{\hat s}_{i,k}}} \right)}}{{\sum\nolimits_{l\in {\cal C}_p} {\exp \left( {{{\hat s}_{i,l}}} \right)} }}, \quad i,k \in {\cal C}_p
\end{equation}
where ${\hat s}_{i,k}$ is a $l_2$-normalized $s_{i,k}$ in terms of $k$.

We assume that two nodes have statistically similar distributions, if those two nodes are close to each other based on this PMF. 
In other words, PMF of a node is designed to serve as a descriptor of the node and, in this sense, we call it Similarity-Based Distribution~(SBD). 
We then employ Jeffreys Divergence~(J-Divergence)~\cite{trove.nla.gov.au/work/11809387} symmetrizing Kullback-Leibler divergence for calculating
the statistical distance between two distributions, as follows:
\begin{equation}
	{D_{J}}\left( {{Q_i}\parallel {Q_j}} \right) = \frac{1}{2}\left( {{D_{KL}}\left( {{Q_i}\parallel {Q_j}} \right) + {D_{KL}}\left( {{Q_j}\parallel {Q_i}} \right)} \right),
	\label{Eq:JDiv}
\end{equation}
where each of $Q_i, Q_j$ is a vector that has $L$ bins for all the other nodes
in ${\cal C}_p$.

Finally, our weight function can be characterized with a Gaussian kernel of the Euclidean distance between two feature vectors ${\bf{f}}_i$ and ${\bf{f}}_j$, and
our statistical distance: 
\begin{equation}
	\textrm{w}({{\bf{f}}_i},{{\bf{f}}_j},{{\bf{S}}_{{C_p}}}) = \exp \left\{ { - \frac{{\left\| {{{\bf{f}}_i} - {{\bf{f}}_j}} \right\|_2^2}}{{2\sigma _d^2}} - \frac{{{D_{J}}\left( {{Q_i}\parallel {Q_j}} \right)}^2}{{2\sigma _r^2}}} \right\},
	\label{eq_SD}
\end{equation}
where hyper parameters, $\sigma _d$ and $\sigma _r$, adjust the degree of
nearness.  
In the Gaussian kernel of our weight function, we call the first term as Euclidean Distance~(ED) and the second term as Statistical Distance~(SD).
In an extreme case of calculating a weight between a node and its hard negative, our SD can identify the hard negative better than ED.
This is because SD sees all the other nodes in the clique, some of which are properly represented in the feature space, even while the hard negative is not.

\noindent \textbf{Multivariate Gaussian form.}
For ease of explanation, we now use a matrix notation within a clique of a specified size $L$ such as a weight matrix,
${\bf{W}}_p \in \mathbb{R}^{L \times L}$, whose element $w_{i,j}$ corresponds to $\textrm{w}({{\bf{f}}_i},{{\bf{f}}_j},{{\bf{S}}_{{{\cal C}_p}}})$.   
From now on, we also treat ${\bf{S}}_{{{\cal C}_p}}$ as a matrix that is composed of elements $s_{i,j}$ for the sake of simple explanation.
A variant~\cite{DBLP:conf/ecai/RadosavljevicVO10} of the C-CRF method simply represents Eq.~\ref{Eq:7} into a multivariate Gaussian form, as follows:
\begin{equation}
	\frac{1}{{{{\left( {2\pi } \right)}^{\frac{L}{2}}}{{\left| \Sigma_p  \right|}^{\frac{1}{2}}}}}\exp \left( { - \frac{1}{2}{{\left( {{{\bf{y}}_p} - {\mu _p}} \right)}^T}{\Sigma_p ^{ - 1}}\left( {{{\bf{y}}_p} - {\mu_p}} \right)} \right).
	\label{eq_Gaussian}
\end{equation}
The covariance matrix $\Sigma_i$ and mean $\mu_i$ is then defined as:
\begin{equation}
	\begin{array}{c}
		{\Sigma_p^{-1}} = 2\left( {\alpha {\bf{I}} + \beta {\bf{D}}_p  - \beta {\bf{W}}_p} \right),\\
		{\mu _p} = \Sigma {\bf{b}}_p ,
	\end{array}
	\label{eq_Gaussian}
\end{equation}
where ${\bf{b}}_p=2\alpha {\bf{s}}_p$, ${{\bf{s}}_p} = {\left[ {{s_{p,1}},...,{s_{p,L}}} \right]^T}$, and $D_p$ is the degree matrix of ${\bf{W}}_p$.

\noindent \textbf{Inference.} 
We can simply find ${\bf{y}}_p^ *$ that maximizes the conditional probability ${\rm{P}_p}\left( {{{\bf{y}}_p}|{{\bf{S}}_{{{\cal C}_p}}}} \right)$ by following the Gaussian property:
\begin{equation}
	{\bf{y}}_p^* = \mathop {\arg \max }\limits_{{{\bf{y}}_p}} {{\rm{P}}_p}\left( {{{\bf{y}}_p}|{{\bf{S}}_{{C_p}}}} \right) = {\mu _p}.
\end{equation}
We then repeat to calculate the solution for the whole database images by setting each database image to a pivot image $I_p$.
We can also employ a conjugate gradient~(CG) method~\cite{shewchuk1994introduction} to calculate the approximated solution of the inverse problem for computational efficiency.
A constraint for CG is satisfied because the covariance matrix is symmetric and positive semi-definite.

\noindent \textbf{Affinity matrix of denoised similarities.}
Existing diffusion methods~\cite{DBLP:conf/cvpr/IscenTAFC17,DBLP:journals/corr/abs-1811-10907} employed the reciprocity check over $k$-NN lists for constructing an affinity matrix due to noises and outliers.
On the other hand, our denoised similarities of $\bf{y}^*$ do not have much of noises and outliers.
Thus we can instead use a simple symmetric affinity from the NN graph of our denoised similarities as $a_{pi} = \left({\bf{y}}_{p,i}^* + {\bf{y}}_{i,p}^*\right)/2$.
Afterward, we follow steps of each of diffusion methods with the denoised, symmetric affinity matrix to perform visual re-ranking.

\vspace{-2mm}
\section{Experiments}
\vspace{-2mm}
\label{sec:experiments} 
In this section, we explore the application of our denoising approach to landmark retrieval with three visual re-ranking methods.
\vspace{-2mm}
\subsection{Experimental setup}
\vspace{-2mm}

\noindent \textbf{Datasets and features.} 
We conduct our experiments using challenging datasets: $\mathcal{R}$Oxford and $\mathcal{R}$Paris~\cite{DBLP:conf/cvpr/RadenovicITAC18} for landmark retrieval. $\mathcal{R}$Oxford and $\mathcal{R}$Paris have three kinds of evaluation protocols, i.e., \textit{Easy}, \textit{Medium}, and \textit{Hard}.
Specifically, $\mathcal{R}$Oxford contains $4,993$ images and $70$ queries that represent particular Oxford landmarks and $\mathcal{R}$Paris consists of $6,322$ images and $70$ queries related to particular Paris landmarks.

We employ GeM networks~\cite{DBLP:journals/pami/RadenovicTC19} of ResNet and VGG versions for landmark retrieval, which are easy to reproduce in a publicly available author's implementation\footnote{\url{https://github.com/filipradenovic/cnnimageretrieval-pytorch}}.

\noindent \textbf{Parameter setup.}
We simply fix C-CRF parameters $\alpha$, $\beta$ to $1$, $0.1$, respectively, for all the following experiments.
Parameters $\sigma _d$, $\sigma_r$ of the Gaussian kernel are set to $0.8$, $2\times10^{-4}$ for VGG and $0.9$, $3.5\times10^{-4}$ for ResNet.
We empirically choose clique sizes of $1,000$ and $500$ for $\mathcal{R}$Oxford and $\mathcal{R}$Paris datasets, respectively.
For other baseline methods, we mainly follow the same parameters of each method.

\vspace{-2mm}
\subsection{Results}
\vspace{-2mm}
\begin{table}
	\caption{mean Average Precision~(mAP) comparisons against other re-ranking methods using the NN-graph for \textbf{$\mathcal{R}$Oxford}. ResNet and VGG for extracting global features are fine-tuned networks. Improvements without online-time cost are indicated in \textcolor{red}{red}, while slight degradations are shown in \textcolor{blue}{blue}.}
    \vspace{-3mm}
	\label{tab:oxford_comparisons}
	\centering
	\fontsize{8}{10}\selectfont
	\renewcommand{\tabcolsep}{2.92mm}
	\renewcommand{\arraystretch}{1.11}
	\begin{tabular}{c| l|ccc}
		\hline
		& Method & \textit{Easy} & \textit{Medium} & \textit{Hard}  \\
		\hline\hline
		\multirow{7}{*}{\rotatebox[origin=c]{90}{ResNet}} & NN-Search & 84.2 & 65.4 & 40.1 \\
		& NNS + AQE & 81.9 & 67.1 & 42.7 \\
		\cline{2-5}
		& EGT & 81.8 & 65.4 & 42.5 \\
		& \textbf{ + Ours} & 85.5\textcolor{red}{\scriptsize(+3.7)} & 73.2\textcolor{red}{\scriptsize(+7.8)} & 50.8\textcolor{red}{\scriptsize(+8.3)}\\
		& Online diffusion & 84.4 & 67.0 & 37.9 \\
		& \textbf{ + Ours} & 91.5\textcolor{red}{\scriptsize(+7.1)} & 73.7\textcolor{red}{\scriptsize(+6.3)} & 45.3\textcolor{red}{\scriptsize(+7.4)}\\
		& Offline adiffusion & 88.2 & 69.9 & 41.1 \\
		& \textbf{ + Ours} & 92.4\textcolor{red}{\scriptsize(+4.2)} & 76.1\textcolor{red}{\scriptsize(+6.2)} & 50.3\textcolor{red}{\scriptsize(+9.2)} \\
		\hline
		
		\hline
		\multirow{7}{*}{\rotatebox[origin=c]{90}{VGG}} & NN-Search & 79.4 & 60.9 & 32. \\
		& NNS + AQE & 86.3 & 69.1 & 41.1 \\
		\cline{2-5}
		& EGT & 88.3 & 71.6 & 44.9 \\
		& \textbf{ + Ours} & 87.5\textcolor{blue}{\scriptsize(-0.8)} & 72.4\textcolor{red}{\scriptsize(+0.8)} & 46.7\textcolor{red}{\scriptsize(+1.8)} \\
		& Online diffusion & 83.2 & 67.4 & 39.9 \\
		& \textbf{ + Ours} & 88.4\textcolor{red}{\scriptsize(+5.2)} & 71.4\textcolor{red}{\scriptsize(+4.0)} & 42.6\textcolor{red}{\scriptsize(+2.7)} \\
		& Offline diffusion & 87.2 & 70.4 & 42.0 \\
		& \textbf{ + Ours} & 90.0\textcolor{red}{\scriptsize(+2.8)} & 73.1\textcolor{red}{\scriptsize(+2.7)} & 45.0\textcolor{red}{\scriptsize(+3.0)} \\
		\hline
	\end{tabular}
    \vspace{-5mm}
\end{table}
\begin{table}
	\caption{Performance (mAP) comparisons against other re-ranking methods using the NN-graph for \textbf{$\mathcal{R}$Paris}. The layout is the same as in Table~\ref{tab:oxford_comparisons}.}
    \vspace{-3mm}
	\label{tab:paris_comparisons}
	\centering
	\fontsize{8}{10}\selectfont
	\renewcommand{\tabcolsep}{2.92mm}
	\renewcommand{\arraystretch}{1.11}
	\begin{tabular}{c| l|ccc}
		\hline
		& Method & \textit{Easy} & \textit{Medium} & \textit{Hard}  \\
		\hline\hline
		\multirow{7}{*}{\rotatebox[origin=c]{90}{ResNet}} & NN-Search & 91.6 & 76.7 & 55.2 \\
		& NNS + AQE & 93.6 & 82.3 & 63.9 \\
		\cline{2-5}
		& EGT & 92.8 & 82.7 & 68.6 \\
		& \textbf{ + Ours} & 92.7\textcolor{blue}{\scriptsize(-0.1)} & 83.3\textcolor{red}{\scriptsize(+0.6)} & 69.6\textcolor{red}{\scriptsize(+1.0)} \\
		& Online diffusion & 93.7 & 88.5 & 78.3 \\
		& \textbf{ + Ours} & 95.0\textcolor{red}{\scriptsize(+2.3)} & 91.0\textcolor{red}{\scriptsize(+2.5)} & 80.8\textcolor{red}{\scriptsize(+2.5)}  \\
		& Offline adiffusion & 94.2 & 87.9 & 77.6 \\
		& \textbf{ + Ours} & 95.1\textcolor{red}{\scriptsize(+0.9)} & 89.3\textcolor{red}{\scriptsize(+1.4)} & 78.7\textcolor{red}{\scriptsize(+1.1)} \\
		\hline
		
		\hline
		\multirow{7}{*}{\rotatebox[origin=c]{90}{VGG}} & NN-Search & 86.8 & 69.3 & 44.2 \\
		& NNS + AQE & 91.0 & 75.4 & 52.7 \\
		\cline{2-5}
		& EGT & 92.3 & 81.2 & 64.7 \\
		& \textbf{ + Ours} & 92.5\textcolor{red}{\scriptsize(+0.2)} & 83.1\textcolor{red}{\scriptsize(+1.9)} & 69.3\textcolor{red}{\scriptsize(+4.6)} \\
		& Online diffusion & 92.7 & 85.9 & 74.2 \\
		& \textbf{ + Ours} & 94.7\textcolor{red}{\scriptsize(+2.0)} & 89.1\textcolor{red}{\scriptsize(+3.2)} & 78.2\textcolor{red}{\scriptsize(+4.0)} \\
		& Offline diffusion & 92.2 & 84.1 & 72.3 \\
		& \textbf{ + Ours} & 93.3\textcolor{red}{\scriptsize(+1.1)} & 85.4\textcolor{red}{\scriptsize(+1.3)} & 73.2\textcolor{red}{\scriptsize(+0.9)} \\
		\hline
	\end{tabular}
    \vspace{-5mm}
\end{table}

\noindent \textbf{Complementarity Analysis.}
For validating the denoising quality, we exploit three re-ranking methods, i.e., online diffusion~\cite{DBLP:conf/cvpr/IscenTAFC17}, offline diffusion~\cite{DBLP:journals/corr/abs-1811-10907}, and Explore-Exploit Graph Traversal~(EGT)~\cite{chang2019explore}.
Tables~\ref{tab:oxford_comparisons}, \ref{tab:paris_comparisons} show performance changes when our method is integrated with other re-ranking methods.
Other tested method include Average Query Expansion~(AQE)~\cite{DBLP:conf/iccv/ChumPSIZ07} on $R$Oxford and $R$Paris datasets.
Overall, the average improvements with our method are 4.6 mAP for $\mathcal{R}$Oxford, 1.7 mAP for $\mathcal{R}$Paris.
These results are mainly acquired thanks to our C-CRF based denoising.

\begin{table}[t]
	\caption{Ablation study of each term constituting our weight function. ``w/ ED'' and ``w/ SD'' represent to use the Euclidean distance of CNN-based features and the statistical distance of our similarity-based descriptors, respectively. Baseline indicates to use only diffusion without C-CRF. The best results are highlighted in \textbf{bold}.
	}
	\vspace{-5mm}
	\centering
	\fontsize{8}{10}\selectfont
	\renewcommand{\tabcolsep}{2.7mm}
	\renewcommand{\arraystretch}{1}
	\label{tab:ablation_study}
	\renewcommand{\arraystretch}{1.15}
	\begin{center}
		\begin{tabular}{c | c c|c c}
			\hline
			 & w/ ED  &w/ SD  &$\mathcal{R}$Oxford& $\mathcal{R}$Paris \\
			\hline\hline
			\multirow{4}{*}{\rotatebox[origin=c]{90}{ResNet}} & & & 69.9 &87.9 \\
			&\checkmark & &75.5 &89.2\\
			& &\checkmark &75.9 &\textbf{89.3}  \\
			&\checkmark &\checkmark &\textbf{76.1} &\textbf{89.3}\\
			\cline{1-5}
			\multirow{4}{*}{\rotatebox[origin=c]{90}{VGG}} & & & 70.4 &84.1\\
			&\checkmark & & 70.9 &84.9  \\
			& &\checkmark &72.7 &85.3   \\
			&\checkmark &\checkmark & \textbf{73.1} &\textbf{85.4}\\
			\hline
		\end{tabular}\vspace{-6mm}
	\end{center}
\end{table}

\noindent \textbf{Ablation study.}
We conduct ablation experiments for each term of the weight function of the C-CRF on offline diffusion~\cite{DBLP:journals/corr/abs-1811-10907}.
The experiments are conducted with the \textit{Medium} protocol of $R$Oxford and $R$Paris, since the \textit{Medium} protocol consisting of both of \textit{Easy} and \textit{Hard} images that can represent an overall quality of retrieval results. 
Table \ref{tab:ablation_study} displays performances with different weight terms on each network for feature extraction.  
As shown in the ablation study, we see that using both terms of the Euclidean Distance~(ED) of CNN features and the Statistical Distance~(SD) of similarity distributions for the weight function yields the best performance.

\begin{figure}
	\centering

	\begin{subfigure}[h]{0.44\columnwidth}
		\includegraphics[width = 1\columnwidth]{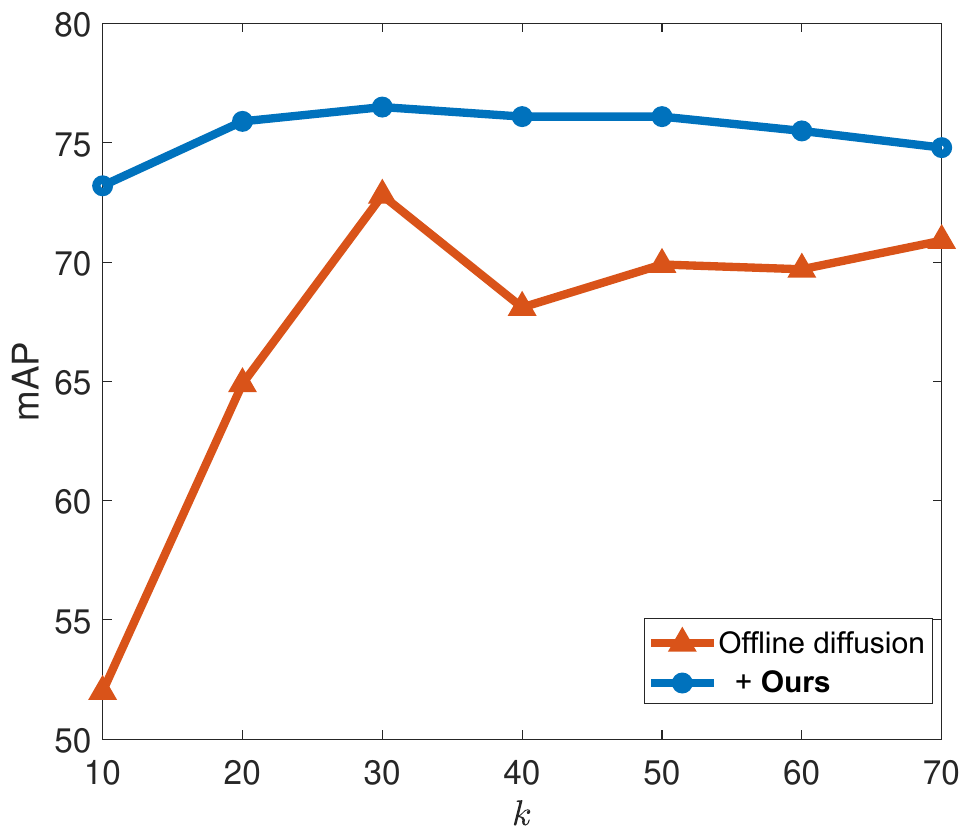}
		\caption{$R$Oxford-ResNet}
	\end{subfigure}
        \hfill
	\begin{subfigure}[h]{0.44\columnwidth}
		\includegraphics[width = 1\columnwidth]{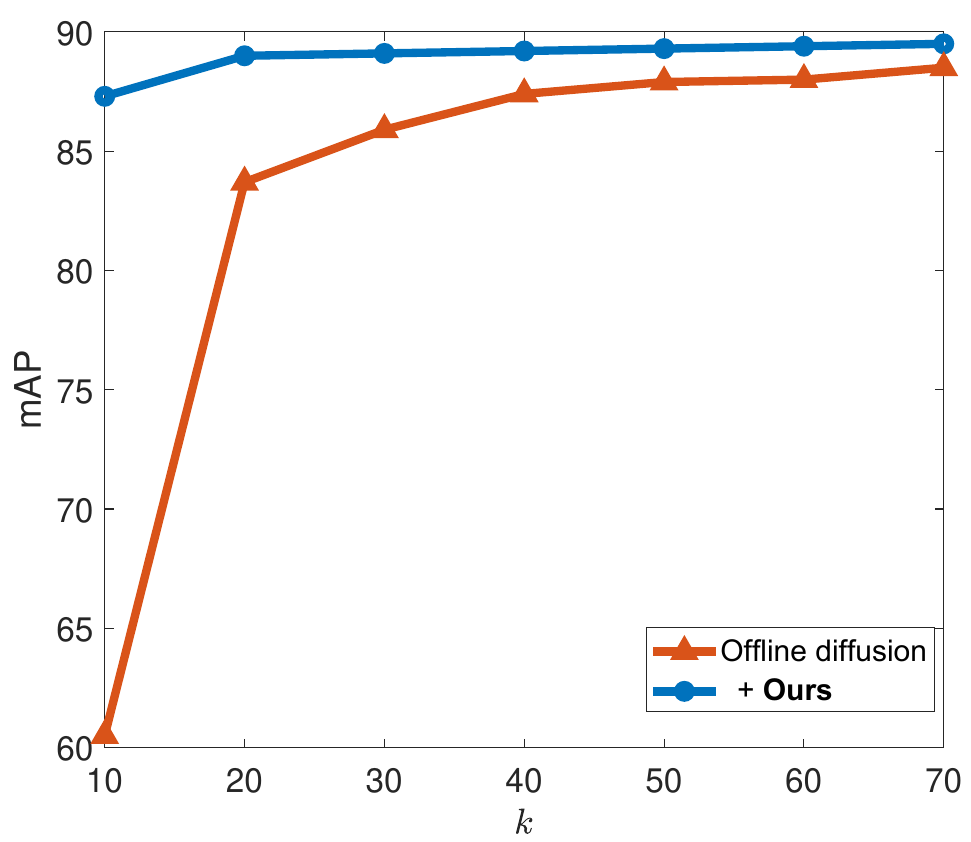}
		\caption{$R$Paris-ResNet}
	\end{subfigure}
        \hfill
        
	\vspace{-3mm}
	\caption{mAP graphs on varying $k$. The mAP results are measured on the \textit{Medium} protocol.}
	\label{fig:varying_k}\vspace{-3mm}
\end{figure}

In Figure~\ref{fig:varying_k}, we additionally experiment with measuring performance effect on varying $k$ parameters since a sparsity of the $k$-NN graph is determined from the $k$, and the sparsity affects the online time complexity; specifically, high density gives rise to high diffusion complexity.
In the same vein, our resulting graph outperforms the reciprocal-NN graph with a more significant gap for smaller k (than 50), and this means we can reduce the online time complexity with less degradation of the retrieval quality. Furthermore, diffusion-based methods are known to be sensitive to the choice of parameters~\cite{magliani2019genetic}. Our approach helps to mitigate the sensitivity to the important parameter $k$.

\begin{figure}
	\centering

	\begin{subfigure}[h]{0.48\columnwidth}
		\includegraphics[width = 1\columnwidth]{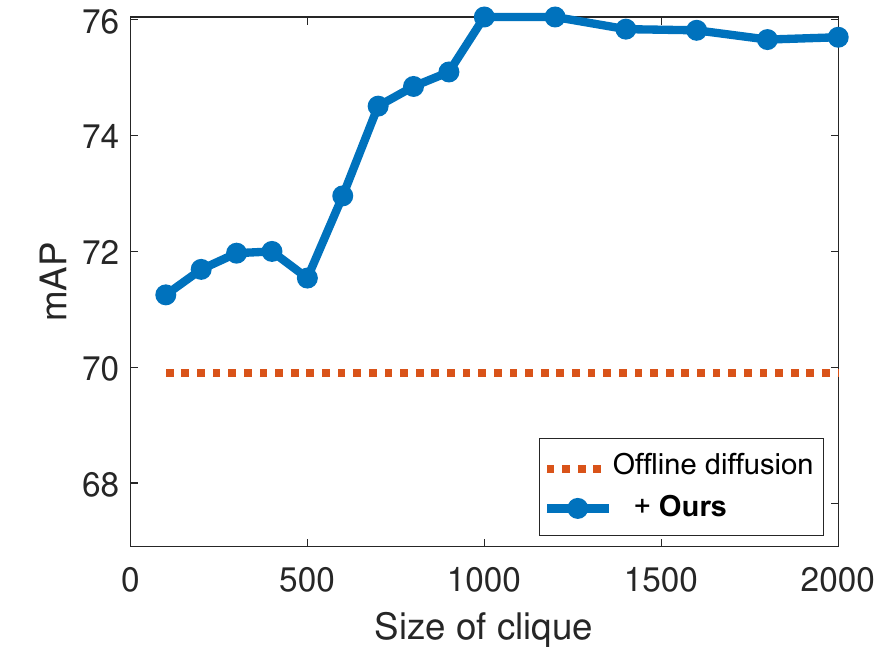}
		\caption{$R$Oxford-ResNet}
	\end{subfigure}%
        \hfill
	\begin{subfigure}[h]{0.48\columnwidth}
		\includegraphics[width = 1\columnwidth]{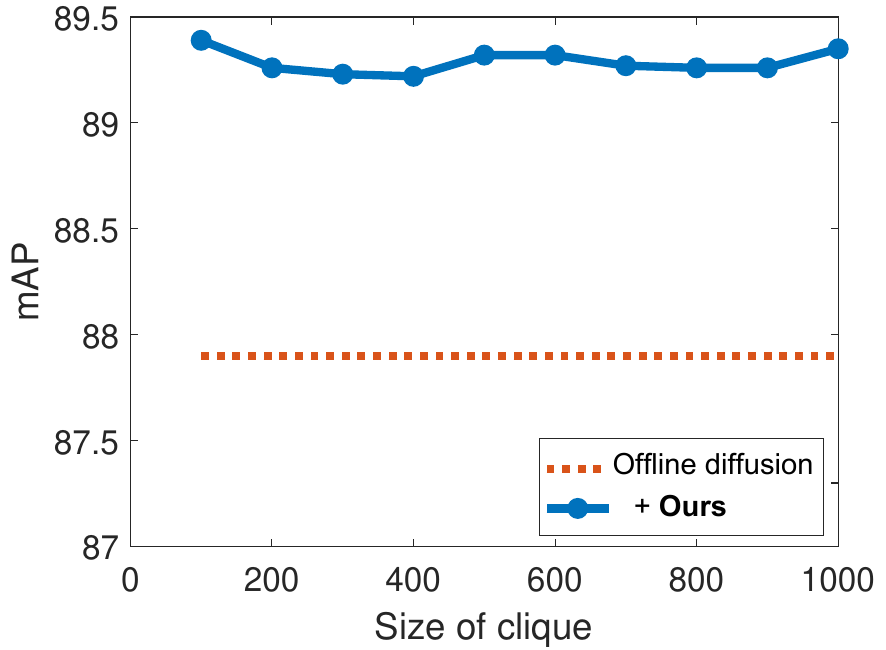}
		\caption{$R$Paris-ResNet}
	\end{subfigure}%
        \hfill
        
	\vspace{-3mm}
	\caption{ mAP graphs on varying sizes of the clique. The mAP results
		are measured on the \textit{Medium} protocol.
	}
	\label{fig:varying_clique}\vspace{-5mm}
\end{figure}

In Figure~\ref{fig:varying_clique}, the mAP performances are shown with varying clique sizes.
The ablation study analyzes the performance according to the clique size, demonstrating robust results over a wide range of clique sizes compared to the baseline. We choose appropriate clique sizes based on this study, even though our method is robust to varying clique sizes in a wide range when compared to baseline.

\vspace{-2mm}
\section{Conclusion}
\vspace{-2mm}
\label{sec:conclusions}
	
We propose a C-CRF-based denoising approach to enhance NN graph-based re-ranking. Our method leverages fully connected cliques and a novel statistical distance metric to robustly alleviate noisy edges. Experimental results demonstrate consistent improvements over existing re-ranking approaches, confirming the effectiveness of our method as a complementary tool for enhancing visual re-ranking.

\vspace{-5mm}
\section*{\normalsize Acknowledgements}
\vspace{-3mm}
This work was supported by the Institute of Information \& communications Technology Planning \& Evaluation~(IITP) grant (No.~RS-2023-00237965, Recognition, Action and Interaction Algorithms for Open-world Robot Service) and the National Research Foundation of Korea~(NRF) grant (No.~RS-2023-00208506), both funded by the Korea government~(MSIT).
Prof. Sung-Eui Yoon is a corresponding author (\texttt{e-mail}: \texttt{sungeui@kaist.edu}).

\bibliographystyle{IEEEbib}
\bibliography{main}
\end{document}